\documentclass[letterpaper, 10 pt, conference]{ieeeconf}  

\IEEEoverridecommandlockouts                              

\usepackage{mathptmx} 
\usepackage{times} 
\usepackage{amsmath} 
\usepackage{amssymb}  
\usepackage{caption}
\usepackage{graphicx}
\usepackage{float}
\usepackage{stfloats}
\usepackage{booktabs}
\usepackage{multicol}
\usepackage{multirow}
\usepackage{threeparttable}
\usepackage{subcaption}
\usepackage{cite}
\usepackage[utf8]{inputenc}
\usepackage[T1]{fontenc}  
\usepackage[hidelinks]{hyperref}
\usepackage[ruled,linesnumbered]{algorithm2e}
\usepackage{color}
\usepackage{svg}

\title{\LARGE \bf
SAGE-ICP: Semantic Information-Assisted ICP
}

\author{Jiaming Cui, Jiming Chen, and Liang Li
\thanks{The authors are with the College of Control Science and Engineering,
        Zhejiang University, Hangzhou, 310027, P. R. China.}
}

\graphicspath{{Images/}}

\begin{document}

\maketitle
\thispagestyle{empty}
\pagestyle{empty}

\begin{abstract}

Robust and accurate pose estimation in unknown environments is an essential part of robotic applications. We focus on LiDAR-based point-to-point ICP combined with effective semantic information. This paper proposes a novel semantic information-assisted ICP method named SAGE-ICP, which leverages semantics in odometry. The semantic information for the whole scan is timely and efficiently extracted by a 3D convolution network, and these point-wise labels are deeply involved in every part of the registration, including semantic voxel downsampling, data association, adaptive local map, and dynamic vehicle removal. Unlike previous semantic-aided approaches, the proposed method can improve localization accuracy in large-scale scenes even if the semantic information has certain errors. Experimental evaluations on KITTI and KITTI-360 show that our method outperforms the baseline methods, and improves accuracy while maintaining real-time performance, \textit{i.e.}, runs faster than the sensor frame rate.

\end{abstract}

\section{Introduction}

Real-time and accurate pose estimation is the basis for mobile robots that require navigation in unknown environments autonomously, where visual odometry (VO) \cite{lategahn2014vision}, LiDAR odometry (LO) \cite{c21} and multi-sensor fusion odometry \cite{campos2021orb} have received much attention. Cameras are already widely used in commercialized autonomous driving systems due to their low cost, but they are susceptible to performance deterioration caused by illumination changes and scale uncertainty. LiDAR can obtain accurate distance measurements directly and is more robust to such changes. With the recent availability of long-range, low-cost, and high-resolution Solid-State-LiDAR, it has become more suitable and affordable for applying LiDAR-based localization systems in real large-scale scenes.

With the development of neural networks, semantic information of the mapped area can be acquired from LiDAR point clouds directly \cite{c1,c2,c3,c4}, which has been used to assist SLAM like generating high-precision semantic maps \cite{c5}, improving pose estimation accuracy \cite{c6,c7} or participating in loop closure detection \cite{c8,c9}. However, it is difficult to be real-time due to the high time complexity of semantic segmentation and complex iterative pose optimization process. Moreover, the accuracy of most LiDAR-based semantic methods depends largely on point cloud semantic segmentation results \cite{c6}, which means that lots of effort is required to label the data and train the model for different environments.

\begin{figure}[htbp]
    \centering
\subcaptionbox{Before}{
	\begin{minipage}[b]{0.24\textwidth}
\centering
    \includegraphics
    [width=\textwidth, height=1.2\textwidth]{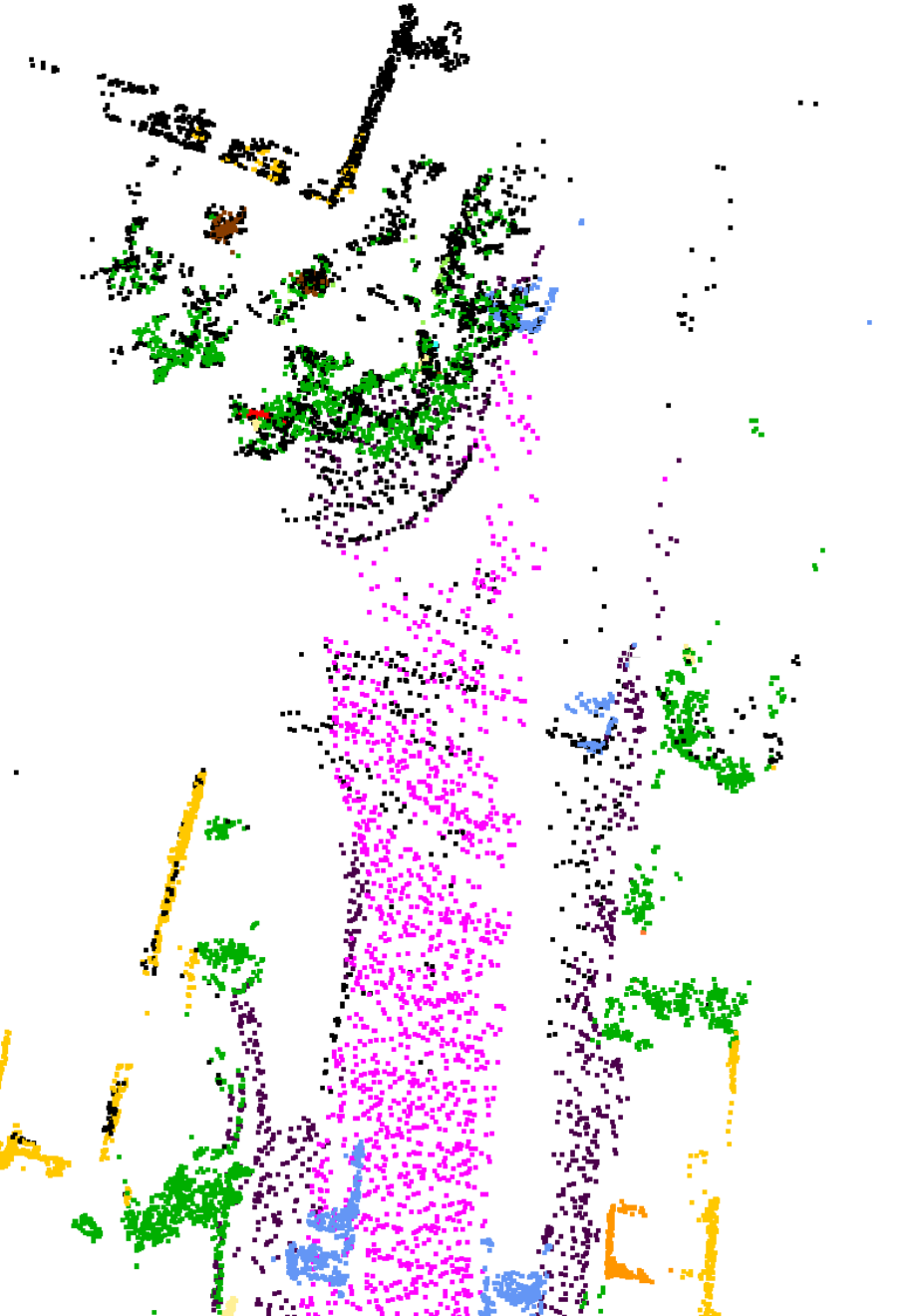}
  \end{minipage}
}
\hspace{-18pt}
\subcaptionbox{After}{
	\begin{minipage}[b]{0.24\textwidth}
\centering
    \includegraphics
    [width=\textwidth, height=1.2\textwidth]{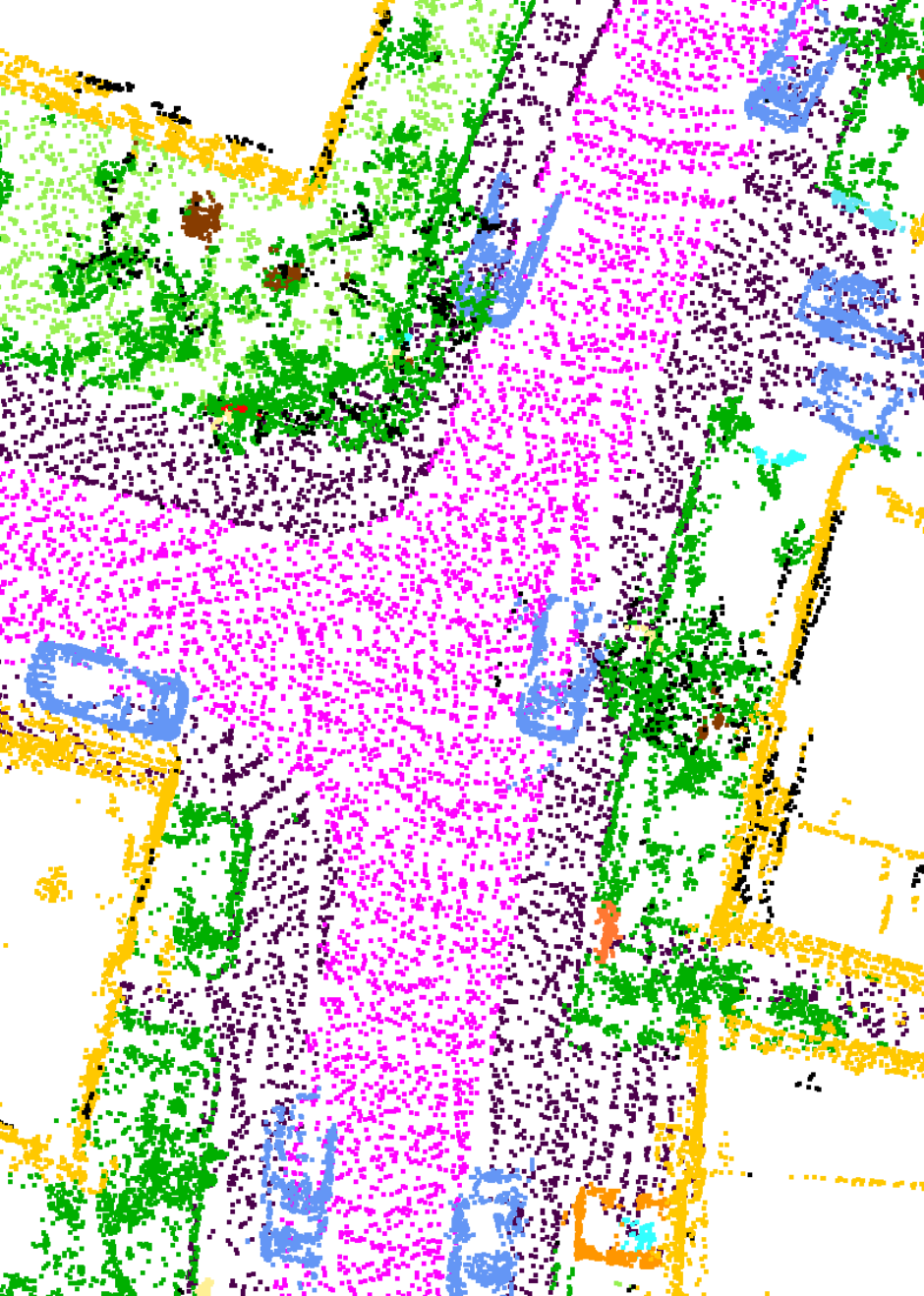}
  \end{minipage}
}
    
    \caption{An example of our adaptive voxel map. (a) and (b) show how our local map is updated before and after the mobile robot passes through the intersection. Points without semantic information (marked in black) are gradually replaced with semantic points.}
    \label{localmap}
\end{figure}


To tackle the above issues, we present SAGE-ICP, a novel semantic information-assisted ICP. Our work is inspired by the idea of KISS-ICP \cite{c10}, which ``keeps it small and simple''. The original point-to-point ICP \cite{c11} requires correctly correlated point pairs and relatively accurate initial values which are not always possible. KISS-ICP improves the performance of ICP but the semantic information is not incorporated in the registration framework. In this paper, we make reasonable use of semantic information to find pairs of correlation points on top of the ICP framework. Nevertheless, whether the points from some objects are dynamic or static cannot be determined only by their semantic labels, \textit{e.g.}, if the car is moving on the road or parked. Filtering out all the possible dynamic points may lead to worse performance of LO as pointed out in \cite{c5}. To deal with this problem, we use prior knowledge to distinguish between dynamic and static vehicles. Instead of gradually removing dynamic objects when updating the map, we remove the relevant point cloud directly before pose estimation, so they have less impact on localization accuracy. On the other hand, the raw point clouds are usually downsampled to reduce the computational time of LO. However, there may exist small objects containing useful information being filtered out in this manner. To solve this problem, different voxel sizes are used for downsampling in this paper and the local map has adaptive storage density for various semantic point clouds.

Inherited from KISS-ICP, the theme of our work is ``simplicity always gives enlightenment''\footnote{The code will be available at \href{https://github.com/NeSC-IV/sage-icp}{https://github.com/NeSC-IV/sage-icp}\iffalse The code will be available upon acceptance of this paper.\fi}. Fig. \ref{localmap} shows an example of our semantic mapping result. The main contributions are summarized as follows:

\begin{itemize}

\item We propose SAGE-ICP, a semantic information-assisted point-to-point ICP. It reserves the small and simple characteristics of KISS-ICP, and makes full use of semantic information to improve accuracy while retaining a high odometry frame rate.
\item We propose an adaptive voxel map with higher storage density for some scarce and critical semantic information.
\item We train the model on the SemanticKITTI \cite{c12,c13}, and extensively evaluate our proposed approach on the KITTI Odometry Benchmark, KITTI road sequences \cite{c14} and KITTI-360 \cite{c15}. The results show that our method outperforms other methods and is robust to erroneous semantic segmentation results.

\end{itemize}

\section{Related Work}

\subsection{Point Cloud Registration}

The mainstream point cloud registration methods can be divided into two categories. The first is ICP-based methods \cite{c10,c11,c16,c17,c18}, which require an initial guess on the transformation and optimize the objective function, \textit{e.g.}, the sum of distances between pairs of associated points, iteratively. ICP-based methods require nearest neighbor (NN) search to find associated point pairs, which is time-consuming. The structure of KD-tree is widely used to speed up the correspondence point pairs association. FAST-LIO2 \cite{c19} proposes an \textit{ikd-Tree} structure that enables incremental map updates, reducing the time for building KD-tree. Moreover, it incorporates IMU to provide an initial pose value and achieves a high convergence speed. Based on \textit{ikd-Tree}, Bai \textit{et al}. \cite{c20} propose a voxel-based LIO algorithm which further speeds up the registration process.

Another type of method is the feature-based approach. They first extract local features, such as planes, lines, and corner points, which are then used to estimate the transformation matrix. Zhang \textit{et al}. \cite{c21} propose LOAM that aligns only planar and edge features to a sparse feature map, which reduces the number of points used for registration. Several other studies \cite{c22,c23,c24} have been built upon the LOAM framework with the goal of enhancing both accuracy and speed, resulting in advancements tailored to specific scenarios. However, these methods are sensitive to noise and incorrect matches. Moreover, careful and tedious parameter tuning is required depending on sensor resolution, experimental scenes, \textit{etc}.

Recent new approaches focus on the runtime and accuracy of the system. CT-ICP \cite{c16} incorporates the motion undistortion into the registration and shows great performance in several datasets. It's noteworthy, however, that CT-ICP necessitates a priori knowledge of the robot's motion model. In contrast, KISS-ICP \cite{c10} gets rid of the necessity for sophisticated optimization techniques to address motion distortion, performing effectively with only the constant velocity model. This strategy enables it to run on different types of equipment without the need to fine-tune the system to a specific application.

\subsection{Semantic-Assisted LiDAR SLAM}

The development of point cloud semantic segmentation algorithms has made it possible to integrate semantic information in pure LiDAR-based SLAM, particularly for large-scale mapping and localization \cite{c5,c25}. SLOAM \cite{c25} is designed specifically for forest inventory. SuMa++ \cite{c5} filters dynamic objects from a surfel-based map, extending ICP with semantic constraints. Some recent studies have also utilized semantic information for loop closure detection \cite{c8,c26,c27}. RINet \cite{c26} addresses place recognition via a global descriptor combining semantic and geometric information. SA-LOAM \cite{c8} integrates a semantic-based loop closure detection method in LOAM, and PADLoC \cite{c27} uses transformer-based panoptic attention for deep loop closure detection. In our approach, we focus on utilizing semantic information in LiDAR odometry. It is important to note that pose graph optimization and loop closure detection can be easily integrated into our approach.

 \begin{figure*}[t]
      \centering
      \includegraphics
      [width=\linewidth]{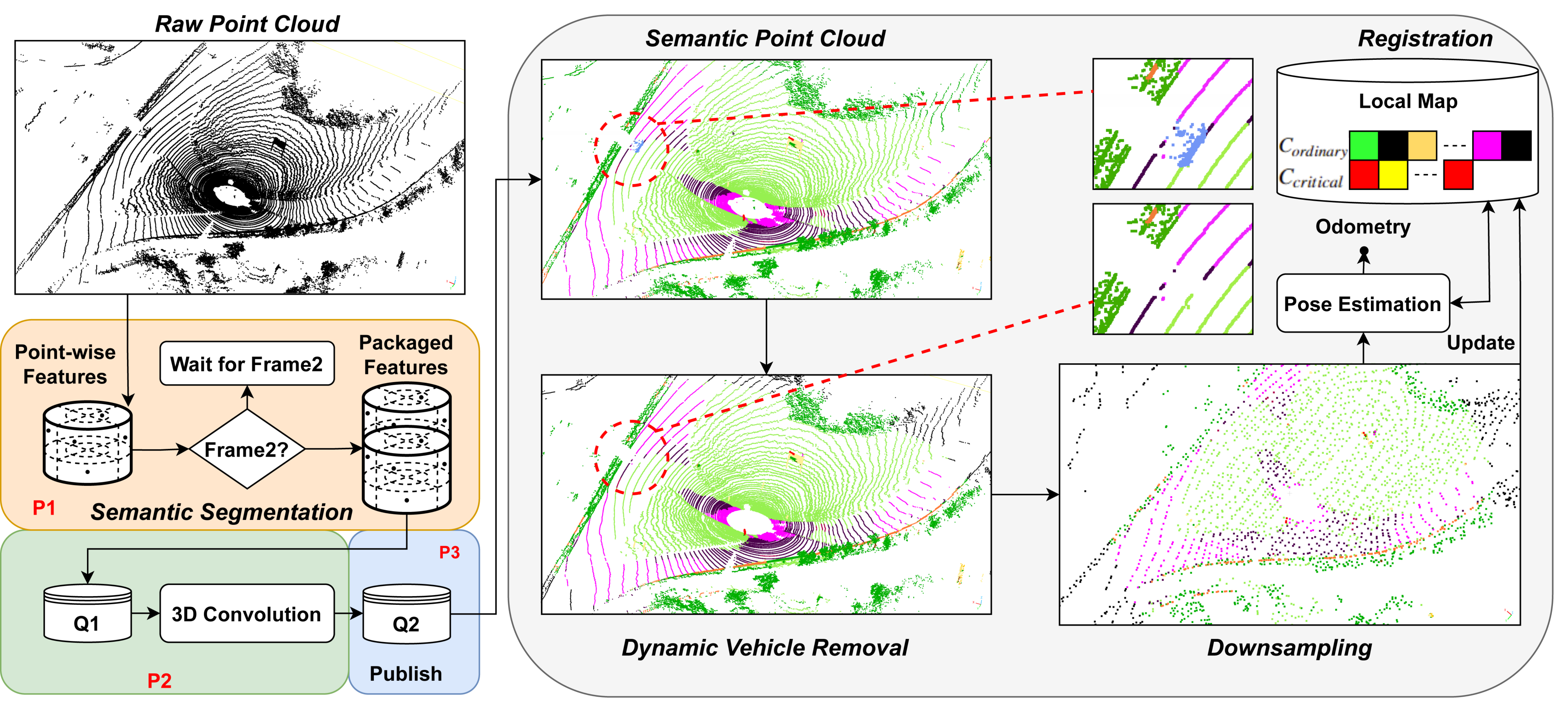}
      \caption{Pipeline of our approach. We perform semantic segmentation in three separate processes to achieve real-time operation in the ROS2 environment.}
      \label{fig2}
   \end{figure*}

\section{SAGE-ICP - Simplicity Always Gives Enlightenment}

This paper aims to leverage semantic information in registration. Fig. \ref{fig2} shows the pipeline of our approach. Firstly, point-wise semantic labels are assigned to the raw point cloud through a point cloud semantic segmentation network \cite{c2}. Then the semantic points belonging to vehicles are further processed to obtain instance segmentation results, while the potential dynamic vehicles are removed using prior knowledge. Subsequently, the point cloud is downsampled individually based on their respective semantic categories, thereby ensuring diverse semantic categories within the downsampled point cloud. The next step involves the alignment of the preprocessed point cloud with the local map. This alignment employs an adaptive threshold for data association \cite{c10}, where the selection of associated points takes into account both the associativity of semantic labels and the Euclidean distance between points. Finally, the adaptive local voxel map is updated, enhancing its capability to store and update more critical semantic points. A comprehensive elaboration of these sequential procedures is provided in the subsequent sections.

\subsection{Real-time Semantic Segmentation}\label{seg}

Cylinder3D \cite{c2} is a cylindrical partition and asymmetrical 3D convolution network which combines point-wise features and voxel-wise inference results together, outperforming range image-based 2D convolution methods \cite{c1,c5}. While the original Cylinder3D is hard to be real-time on a laptop for robotic applications, Hou {et al}. \cite{c4} propose a point-to-voxel knowledge distillation approach named PVD, significantly speeding up the Cylinder3D model and performing best in pure LiDAR semantic segmentation on the SemanticKITTI \cite{c12}. The inference time of the original Cylinder3D model is 170 ms and PVD is 76 ms, but the code of knowledge distillation is not yet open-sourced.

Building upon the original Cylinder3D model, we propose a novel method capable of achieving real-time semantic segmentation within the ROS2 framework\cite{c28} on a laptop (see Fig. \ref{fig2}). Our method consists of two key modifications to the point cloud segmentation procedure. First, we split the entire semantic labels inference process into three distinct stages; Second, we merge pairs of successive frames to be fed into the neural network. The first frame of the raw point cloud is first rasterized by the cylindrical partition and point-wise features are extracted by MLP. To ensure the real-time performance of process 1 (P1 in Fig. \ref{fig2}), we randomly select a point in each voxel to extract features as the voxel-wise features. The second frame is treated in the same way and then packaged into queue 1 along with the first frame. Asymmetrical 3D convolution networks are then used to generate the voxel-wise outputs, while all points within the same voxel are considered to have identical labels (P2 in Fig. \ref{fig2}). Two frames of point clouds with semantic information are pushed back sequentially into queue 2, while process 3 publishes them in turn (P3 in Fig. \ref{fig2}). Since the three processes are independent of each other, and the time consumed by process 2 to perform an inference is less than 200ms, therefore, if process 3 publishes a frame of semantic point cloud every 100ms, the frequency of the semantic point cloud is the same as that of the raw, with a fixed delay of 300ms. A detailed analysis of time consumption is shown in Table \ref {time}.

\subsection{Semantic Point Cloud Preprocessing}

Our preprocessing of the semantic point cloud obtained from Section \ref{seg} consists of two parts: dynamic points removal and class-wise downsampling, to achieve faster convergence, higher robustness, and more accurate registration results.

\noindent\textbf{Removing Dynamic Instance.}\label{rdi} Real-world scenes often have a large number of moving objects. Traditional geometric feature-based approaches suffer from environmental changes, causing significant registration errors. SuMa++ \cite{c5} filters dynamic surfels by checking semantic consistency when a new frame is updated to the map, and has shown that directly removing all dynamic semantic categories leads to a degeneration in odometry accuracy. As a result, it does not remove moving objects before the point cloud registration, which means that dynamics still affect pose optimization to some extent. Since the positions of static vehicles in real scenarios have certain regularity (\textit{e.g.}, on parking lots, by the sidewalk), we propose a simple yet effective approach to distinguish dynamic and static vehicles using prior knowledge. Fig. \ref{fig2} shows an example before and after filtering dynamics. We first categorize semantic points into two distinct groups: dynamic ${\mathbf{C_d}}$ and static ${\mathbf{C_s}}$. Subsequently, we utilize Euclidean point cloud clustering from the PCL library \cite{c29} to effectively distinguish various instances within ${\mathbf{C_d}}$ denoted as ${\mathbf{I_d}}$. Then we deal with each instance ${i_{d}^k} \in \mathbf{I_d}$ individually. For each instance, we search for a specific range ${d_r}$ within the point cloud ${\mathbf{P_{r}^k}}$ surrounding ${i_{d}^k}$. This range is identified using a KD-tree constructed from the static points within $\mathbf{{C_s}}$. We then tally the count of landmark points ${\mathbf{P_{ic}^k}}$ falling within this range. If the value of ${\theta _{ic}}$ surpasses a designated threshold ${\theta_{0}}$, we suppose that the instance ${i_{d}^k}$ is stationary:

\begin{equation}
{\theta _{ic}} = \frac{{|\mathbf{P_{ic}^k}|}}{{|\mathbf{P_r^k}|}}
\end{equation}

\noindent where ${|\mathbf{P^k}|}$ is the number of points in ${\mathbf{P^k}}$.

\noindent\textbf{Semantic Subsampling.}\label{ss} Extracting keypoints from dense raw point clouds is an essential step for feature-based scan registration \cite{c21,c22,c23,c24}, which usually has many parameters to be tuned for different scenes. Since semantic information is an advanced environmental feature, complex geometric feature extraction can be avoided. We adopt point cloud subsampling approach from KISS-ICP \cite{c10} to maintain one point per voxel in the original coordinates. Moreover, we further use different voxel grid sizes for different categories, to prevent some small but critical points (\textit{e.g.}, road signs, lane lines) inevitably from being filtered out.

\subsection{Semantic Assisted Association}\label{saa} Data association is a prerequisite for iterative pose optimization, \textit{i.e.}, finding the correct point associations. Most ICP-based methods adopt the standard nearest neighbor approach to handle this task, while semantic methods \cite{c5,c6,c7} directly associate two points with the same label, or give a weight coefficient according to whether the semantic label of the associated point pair is consistent.

As the semantic segmentation result cannot be 100$\%$ correct, our strategy is to integrate semantic labels into the nearest neighbor search and avoid associating points from different categories. For each source point in the preprocessed point cloud ${p_s^k} \in \mathbf{P_{pre}}$, we firstly calculate its voxel coordinates based on motion prediction ${T_{pred,t}}$, and collect all points $\mathbf{P_{ls}^k}$ within that voxel and neighboring voxels from the local map. The semantic Euclidean distance ${d_{se,j}}$ of each point ${p_{ls,j}^k} \in \mathbf{P_{ls}^k}$ from the source point is defined as

\begin{equation}
{d_{se,j}} = {\gamma{||p_{ls,j}^k-p_s^k||}_2}
\end{equation}

\noindent where semantic correlation coefficient $\gamma$ is given by

\begin{equation}
\gamma =
\begin{cases}
\gamma_0,  & \text{if $l(p_{ls,j}^k)=l(p_s^k)$ or $l(p_{ls,j}^k)l(p_s^k)=0$} \\
1, & \text{otherwise}
\end{cases}
\end{equation}

\noindent  where $l(p_{ls,j}^k)$, $l(p_s^k)$ are semantic labels of ${p_{ls,j}^k}$ and $p_s^k$, respectively, and $\gamma_0$ is a constant in range $(0,1)$. $l\left(\cdot\right)=0$ indicates that the point has no semantic category. Since the inference error of Cylinder3D \cite{c2} increases with distance, we manually remove labels from point cloud 50 meters away from the center of the LiDAR. Then we get a target point ${p_{ls,{j0}}^k} \in \mathbf{P_{ls}^k}$ with the smallest semantic Euclidean distance ${d_{se,j0}}$ and store this point pair in $\mathbf{\mathcal{C}}\left(\tau_t\right)$ if their Euclidean distance (not ${d_{se,j0}}$) is below adaptive threshold $\tau_t$ computed by constant velocity model

\begin{equation}
\begin{aligned}
 & \tau_t=3\sigma_t, \\
 & \sigma_t=\sqrt{\frac{1}{\left|\mathbf{M}_t\right|} \sum_{i \in \mathbf{M}_t} \delta\left(\Delta T_i\right)^2}, \\
 & \mathbf{M}_t=\left\{i \mid i<t \wedge \delta\left(\Delta T_i\right)>\delta_{\min }\right\}
\end{aligned}
\end{equation}
\noindent  where $\Delta T_i$ represents how much deviation there is between odometry and motion prediction, $\delta\left(\Delta T_i\right)$ is maximum point displacement, see \cite{c10} for detail.

Finally, we perform a robust optimization minimizing residuals and repeat an iterative process similar to other ICP methods.




\begin{algorithm}[t]
\caption{Adaptive Voxel Map}\label{algorithm1}
\SetKwInOut{Input}{Input}\SetKwInOut{Output}{Output}
\Input{current point cloud in map coordinate $\mathbf{P_{pre,t}^*}$, current pose $p_t$, local map $M_{t-1}$}
\Output{updated local map $M_t$}
\ForEach{semantic point $p_i^* \in \mathbf{P_{pre,t}^*}$}{%
    $v_i^* \in \mathbb{Z}^3\leftarrow \text{Get voxel grid index from }p_i^*$\;
      $n_i \gets \text{The number of points already in }\mathbf{P(v_i^*)}$\;
      \lIf{$n \textless N_1$}{Add $p_i^*$ to $\mathbf{P(v_i^*)}$}
      \ElseIf{$n \geq N_1 \& n \leq N_2$}{
      \uIf{$l(p_i^*) \in \mathbf{C_{ordinary}}$}{
        \ForEach{$p_{i,j} \in \mathbf{P(v_i^*)}$}{
        \If{$l(p_{i,j})=0$}{Replace $p_{i,j}$ with $p_i^*$\;}
        }
      }
      \lElseIf{$l(p_i^*) \in \mathbf{C_{critical}}$}{Add $p_i^*$ to $\mathbf{P(v_i^*)}$}
      }
    }
\ForEach{non-empty $\mathbf{P(v_k)} \in M_{t-1}$}{
\If{$||\mathbf{P(v_k)_{center}}-p_t||_2 \geq r_{max}$}{Remove $\mathbf{P(v_k)}$}
    }

\end{algorithm}

\subsection{Adaptive Voxel Map}\label{avm}

The widely used approaches to store local maps are voxel grids \cite{c21}, surfels \cite{c5} or \textit{ikd-Tree} \cite{c19}. Hash table is used to store voxels in recent works \cite{c10,c20}, allowing memory-efficient map storage and faster nearest neighbor search. Our adaptive voxel map is based on \cite{c10}, with higher storage density for some critical semantic points and the ability to update point cloud semantic labels (see Algorithm \ref{algorithm1}). When the number of points in the voxels is in the range of $\left[N_1,N_2\right]$, the ordinary points $p_i^* \in \mathbf{C_{ordinary}}$ are used to replace the unlabeled points that have been stored, while the critical points $p_i^* \in \mathbf{C_{critical}}$ are stored as usual. In this way, we can guarantee that our adaptive voxel map contains richer semantic information than other approaches. We also remove voxels that are too far away from the current pose $p_t$ to save memory usage.

\begin{table*}[h]
\caption{Parameters of SAGE-ICP}
\centering
\label{param}
\begin{threeparttable}
\begin{tabular}{c|ccccccccccccc}
\toprule
Datasets&$S_{road}$&$S_{plant}$&$S_{object}$&$S_{vehicle}$&$S_{building}$&$S_{unlabel}$&$\gamma_0$&${\theta _0}$&$N_1$&$N_2$&$r_{max}$&$\tau_0$&$\delta_{\min }$ \\
\midrule
KITTI Odometry&0.6m&0.9m&0.8m&0.6m&1.0m&\multirow{2}{*}{1.0m}&0.4&\multirow{2}{*}{0.5}&\multirow{2}{*}{20}&\multirow{2}{*}{40}&\multirow{2}{*}{100m}&\multirow{2}{*}{2m}&\multirow{2}{*}{0.1m} \\
KITTI Raw/KITTI-360&1.0m&1.0m&0.5m&0.5m&0.5m&&0.8&&&&&& \\
\bottomrule
\end{tabular}
\begin{tablenotes}
    \footnotesize
    \item $S_{road}$ is downsampling grid size of road, parking, sidewalk and other ground; $S_{plant}$ stands for vegetation and terrain; $S_{object}$ stands for trunks, lane-markings, poles, traffic signs and other objects, small but critical points mentioned in Section \ref{ss}; $S_{building}$ stands for buildings and fences.
     \item Due to the different semantic segmentation accuracy of different datasets, we select appropriate parameters for three datasets separately.
\end{tablenotes}
\end{threeparttable}

\end{table*}

\begin{table*}[h]
\caption{Relative Pose Error on KITTI and KITTI-360}
\centering
\label{kitti}
\resizebox{\textwidth}{!}{
\begin{threeparttable}
\begin{tabular}{c|cccccccccccc}
\toprule
KITTI Odometry&00&01&02&03&04&05&06&07&08&09&10&Avg. \\
\midrule
LOAM \cite{c21} &\multicolumn{1}{l}{0.78/-}&\multicolumn{1}{l}{1.43/-}&\multicolumn{1}{l}{0.92/-}&\multicolumn{1}{l}{0.86/-}&\multicolumn{1}{l}{0.71/-}&\multicolumn{1}{l}{0.57/-}&\multicolumn{1}{l}{0.65/-}&\multicolumn{1}{l}{0.63/-}&\multicolumn{1}{l}{1.12/-}&\multicolumn{1}{l}{0.77/-}&\multicolumn{1}{l}{0.79/-}&\multicolumn{1}{l}{0.84/-} \\
KISS-ICP\cite{c10} &0.51/0.19&0.72/\pmb{0.11}&0.52/0.15&0.66/\pmb{0.16}&0.35/0.14&0.30/0.14&0.26/\pmb{0.08}&\pmb{0.32}/0.17&0.82/\pmb{0.18}&0.49/\pmb{0.13}&0.57/0.19&0.50/0.15 \\
SuMa++$^\star$\cite{c5}&0.65/0.22&1.63/0.47&3.54/\pmb{0.14}&0.67/0.47&0.34/0.27&0.40/0.19&0.47/0.27&0.39/0.28&1.01/0.34&0.58/0.20&0.67/0.30&0.94/0.28 \\
SA-LOAM$^{\circ\star}$\cite{c8}&0.59/\pmb{0.18}&1.89/0.48&0.79/0.27&0.87/0.46&0.59/0.35&0.37/0.16&0.52/0.24&0.41/0.18&0.84/0.25&0.77/0.25&0.78/0.35&0.76/0.28 \\
Ours-RangeNet&0.51/0.19&0.65/\pmb{0.11}&0.50/0.15&0.66/\pmb{0.16}&0.33/\pmb{0.12}&0.28/\pmb{0.13}&\pmb{0.25}/0.09&0.33/\pmb{0.15}&\pmb{0.80}/\pmb{0.18}&\pmb{0.47}/0.14&0.55/0.17&\pmb{0.48}/0.15 \\
Ours-Cylinder3D&0.51/0.19&0.67/\pmb{0.11}&0.51/0.16&\pmb{0.64}/0.17&0.33/0.13&0.29/0.14&0.26/0.09&\pmb{0.32}/\pmb{0.15}&0.81/\pmb{0.18}&\pmb{0.47}/0.14&0.55/0.17&0.49/0.15 \\
Ours-ST&\pmb{0.50}/0.19&\pmb{0.64}/\pmb{0.11}&\pmb{0.49}/0.15&0.67/\pmb{0.16}&\pmb{0.32}/0.14&\pmb{0.27}/\pmb{0.13}&\pmb{0.25}/\pmb{0.08}&\pmb{0.32}/\pmb{0.15}&\pmb{0.80}/\pmb{0.18}&0.48/0.14&\pmb{0.51}/\pmb{0.16}&\pmb{0.48}/\pmb{0.14} \\
\bottomrule
\toprule
KITTI-360&00&-&02&03&04&05&06&07&-&09&10&Avg. \\
\midrule
KISS-ICP\cite{c10} &0.73/0.33&&\pmb{0.61}/\pmb{0.30}&\pmb{0.58}/\pmb{0.20}&\pmb{0.75}/\pmb{0.35}&0.71/0.43&0.89/0.38&0.51/\pmb{0.15}&&0.78/\pmb{0.31}&\pmb{0.70}/\pmb{0.25}&\pmb{0.69}/0.30 \\
Ours-RangeNet&0.72/0.33&&\pmb{0.61}/0.31&0.65/0.22&0.79/0.36&\pmb{0.68}/0.42&\pmb{0.67}/\pmb{0.31}&0.55/0.17&&0.78/\pmb{0.31}&0.79/0.27&\pmb{0.69}/0.30 \\
Ours-Cylinder3D&\pmb{0.67}/\pmb{0.30}&&0.62/0.31&0.62/0.21&0.77/\pmb{0.35}&0.69/\pmb{0.40}&0.74/0.34&\pmb{0.50}/\pmb{0.15}&&\pmb{0.77}/0.32&0.77/0.27&\pmb{0.69}/\pmb{0.29} \\
\bottomrule
\toprule
KITTI-PART&30&31&32&33&34&Avg1&&t0&t1&t2&t3&Avg2 \\
\midrule
KISS-ICP\cite{c10} &0.56/\pmb{0.13}&1.75/0.59&1.56/0.43&1.41/\pmb{1.30}&\pmb{0.93}/0.16&1.24/0.52&&0.33/\pmb{0.22}&\pmb{0.33}/\pmb{0.18}&0.94/0.44&0.74/0.38&0.59/0.31 \\
Ours-RangeNet&0.55/0.16&2.18/1.04&2.28/0.72&\pmb{1.39}/\pmb{1.30}&0.95/\pmb{0.15}&1.47/0.67&&\pmb{0.32}/0.24&0.38/0.25&0.88/0.41&\pmb{0.71}/0.40&0.57/0.32 \\
Ours-Cylinder3D&\pmb{0.51}/\pmb{0.13}&\pmb{1.67}/\pmb{0.50}&\pmb{1.39}/\pmb{0.35}&\pmb{1.39}/\pmb{1.30}&0.95/\pmb{0.15}&\pmb{1.18}/\pmb{0.49}&&0.33/0.25&\pmb{0.33}/0.20&\pmb{0.85}/\pmb{0.39}&0.72/\pmb{0.37}&\pmb{0.56}/\pmb{0.30} \\
\bottomrule
\end{tabular}
\begin{tablenotes}
    \footnotesize
    \item RTE in $\%$ / RRE in degrees per 100m. The best performance for each sequence is marked in bold numbers. Approaches marked with $^\star$ indicate semantic assisted odometry and $^\circ$ contain loop closures. The result of LOAM is from the origin paper \cite{c21} while both SuMa++ and SA-LOAM are from \cite{c8}.
    \item We renamed KITTI road sequences "2011$\_$09$\_$26$\_$drive$\_$0015 - 2011$\_$09$\_$26$\_$drive$\_$0032" into KITTI-PART sequences 30-34, and KITTI-360 test SLAM "00-03" into "t0-t3".
\end{tablenotes}
\end{threeparttable}
}
\end{table*}

\section{Experiments}

We design our experiments to demonstrate that our system (i) outperforms state-of-the-art odometry algorithms in most cases, and (ii) is robust to erroneous semantic segmentation results. All parameters are shown in Table \ref{param}. For all sequences from KITTI and KITTI-360, we correct LiDAR scans for an angle of 0.205$^\circ$ to be set up as \cite{c16}.

We evaluated our approach on a laptop with a 4.70 GHz 14-core Intel i7-12700H CPU with 16GB of RAM and an Nvidia GeForce RTX 3070 Ti with 8 GB RAM. The Cylinder3D \cite{c2} model for point cloud semantic segmentation is trained using all sequences from \cite{c12} and semantic labels are provided for KITTI Odometry Benchmark \cite{c14}. It is noteworthy that we use the same model when testing other datasets, which means that our model inevitably produces greater errors when segmenting these data than KITTI Odometry Benchmark. Moreover, We also evaluated Rangenet++ \cite{c1} to prove that our approach is also suitable for other semantic segmentation models. Both semantic segmentation and odometry part are implemented on ROS2 Humble \cite{c28}.

\subsection{KITTI Odometry Benchmark}

The KITTI Odometry Benchmark \cite{c14} consists of 11 training sequences with LiDAR scans. We test the Relative Translation Error (RTE) in percentage and the Relative Rotational Error (RRE) in degrees per 100 m defined by KITTI, which averages the trajectory errors of length ranging from 100 to 800 m.
We present the results of our approach with semantic segmentation values by Cylinder3D (Ours-Cylinder3D) and Rangenet++ (Ours-Rangenet), and also report the results with ground-truth segmentation values (Ours-ST) respectively. The state-of-the-art approach KISS-ICP \cite{c10} is chosen as our baseline. Besides, we compare our results with LOAM \cite{c21}, SuMa++ \cite{c5} and SA-LOAM \cite{c8} in Table \ref{kitti}. LOAM is the state-of-the-art pure LiDAR-based method on the KITTI benchmark, while SuMa++ and SA-LOAM are semantic-aided LiDAR SLAM systems.

Both KISS-ICP and our approach outperform other methods on most sequences. Compared with KISS-ICP, our method has better performance on each sequence with different margins, especially 01 and 10. These sequences are collected from highways or the countryside, which means that there are more dynamic vehicles and fewer geometric feature points in the raw point clouds. We also find that in other sequences collected from urban, most feature points can be obtained directly by classical methods, \textit{i.e.}, building corner points or planes. Our method is slightly better than the baselines though the superiority of our semantic method is not as obvious as that in the dynamic or featureless environments. Although semantic labels generated by Cylinder3D or Rangenet++ are not completely reliable, the results of Ours-Cylinder3D, Ours-Rangenet++, and Ours-ST show that semantic information with partial errors can also benefit localization. Besides, similar to \cite{c5}, we find that removing all points belonging to vehicles leads to worse performance as parked cars are also a reliable source of feature points in some environments.

\begin{table*}[h]
\caption{Time Consumption Analysis (ms)}
\centering
\label{time}
\begin{threeparttable}
\begin{tabular}{c|c|c|c|c|c|c|c|c}
\toprule
\multicolumn{4}{c|}{Semantic Segmentation}&\multicolumn{5}{c}{Registration} \\
\midrule
MLP-Frame1&MLP-Frame2&conv&publish&RDI\tnote{*}&Downsampling&Optimization&Update&Total \\
\midrule
66.94&94.67&168.42&20.82&30.09&40.13&15.13&0.82&86.17 \\
\bottomrule
\end{tabular}
\begin{tablenotes}
    \footnotesize
    \item[*] Removing Dynamic Instance
\end{tablenotes}
\end{threeparttable}
\end{table*}

\subsection{Generalization Validation}

\begin{figure}[t]
    \centering
    \subcaptionbox{Cylinder3D}{
	\begin{minipage}[b]{0.24\textwidth}
\centering
    \includegraphics
    [width=\textwidth, height=1.3\textwidth]{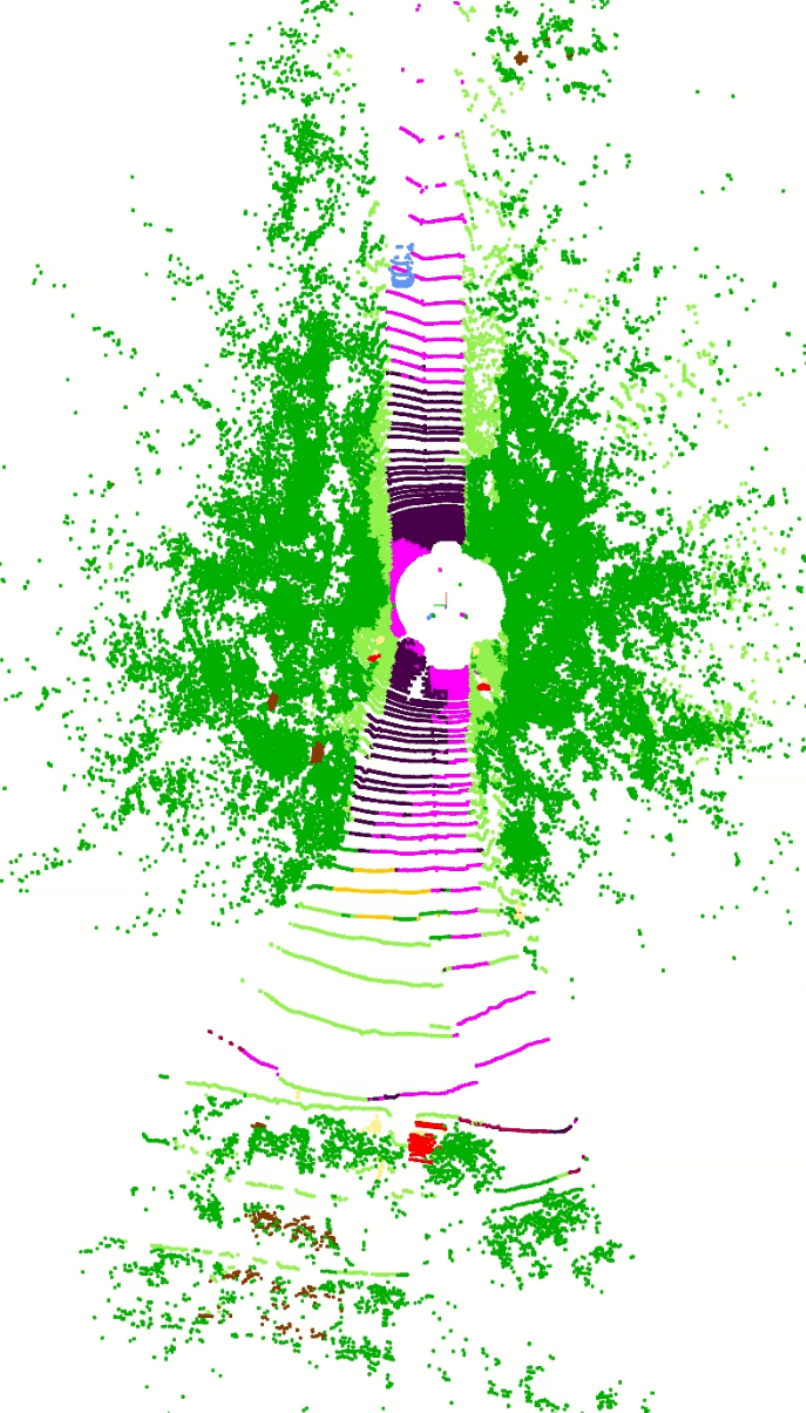}
  \end{minipage}
}
\hspace{-18pt}
    \subcaptionbox{Rangenet++}{
	\begin{minipage}[b]{0.24\textwidth}
\centering
    \includegraphics
    [width=\textwidth, height=1.3\textwidth]{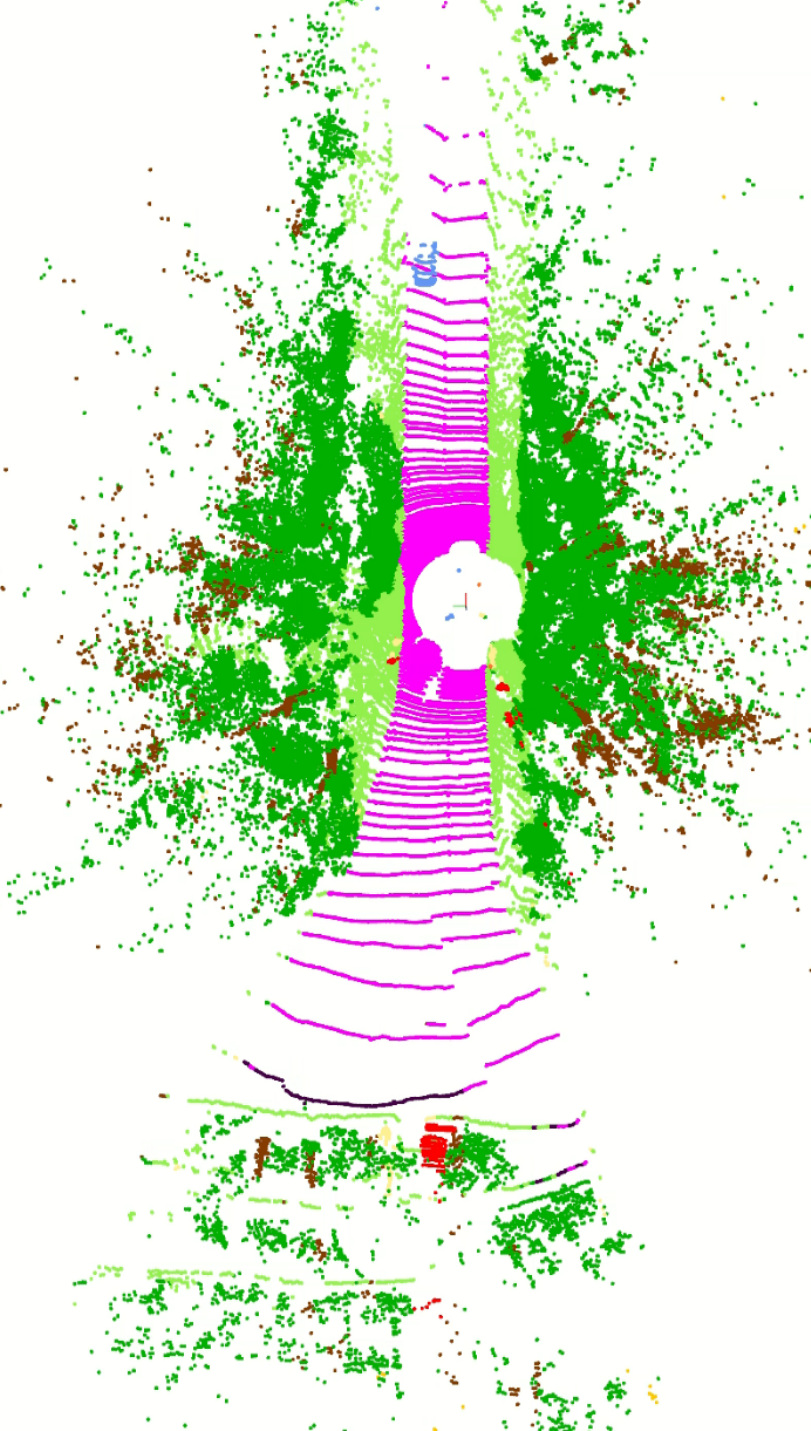}
  \end{minipage}
}
    
    \caption{Semantic segmentation results. (a) Cylinder3D is more accurate for the segmentation of small targets, \textit{e.g.}, trunks (marked with brown), and road signs (marked with red). (b) Rangenet++ is more accurate for semantic segmentation over large areas nearby, \textit{e.g.}, roads (marked with pink).}
    \label{sem_result}
\end{figure}

To further evaluate the generalization ability of our system and the advantages of our approach in dynamic scenes, we compare the same metrics with baseline on KITTI raw data and KITTI-360. The scans in KITTI raw data are not motion-corrected, but the data frame numbers correspond across all sensor streams, and GPS/OXTS data is reliable due to less interference. We selected the first five sequences from the road category collected in the countryside or highway. There are few distinct features in these scenes and flooded with high-speed moving vehicles, which are challenging for SLAM. Results of KITTI-PART in Table \ref{kitti} show that our semantic information-assisted method with the Cylinder3D model achieves more robust and accurate localization performance than KISS-ICP, which suffers from false correspondences on moving objects. Meanwhile, lower rotation error means that our method can capture small but critical feature points from degraded environments such as road signs on the highway. Rangenet++ has poor segmentation accuracy for far and small objects (see Fig. \ref{sem_result}), resulting in large errors.

\begin{figure}[t]
    \centering
    \subcaptionbox{KITTI-360 06}{
	\begin{minipage}[b]{0.5\textwidth}
\centering
\includegraphics[width=0.485\textwidth, height=0.4\textwidth]{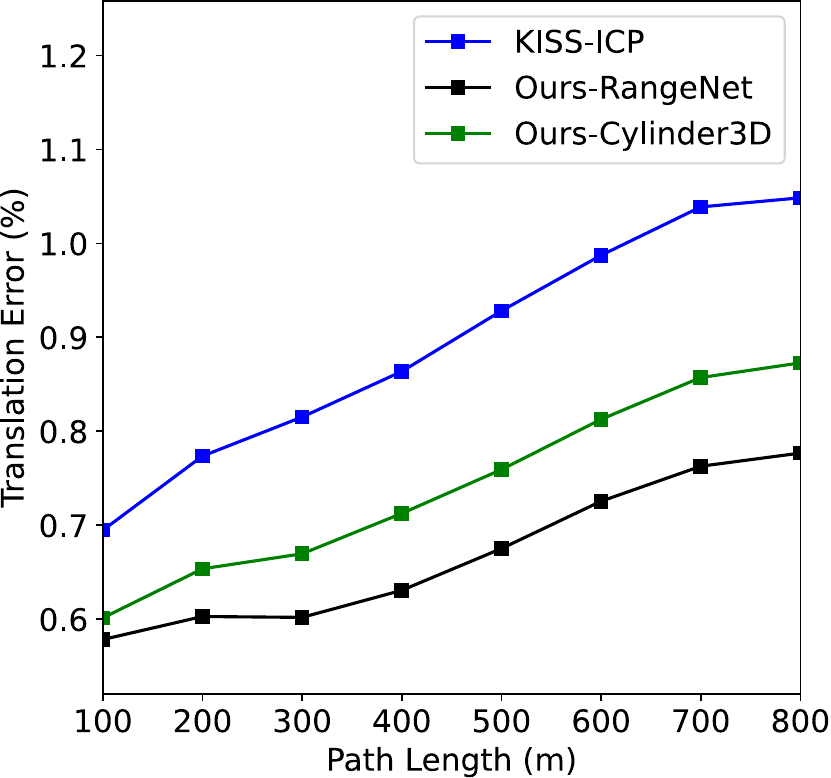}
\hspace{-7pt}
\includegraphics[width=0.487\textwidth, height=0.4\textwidth]{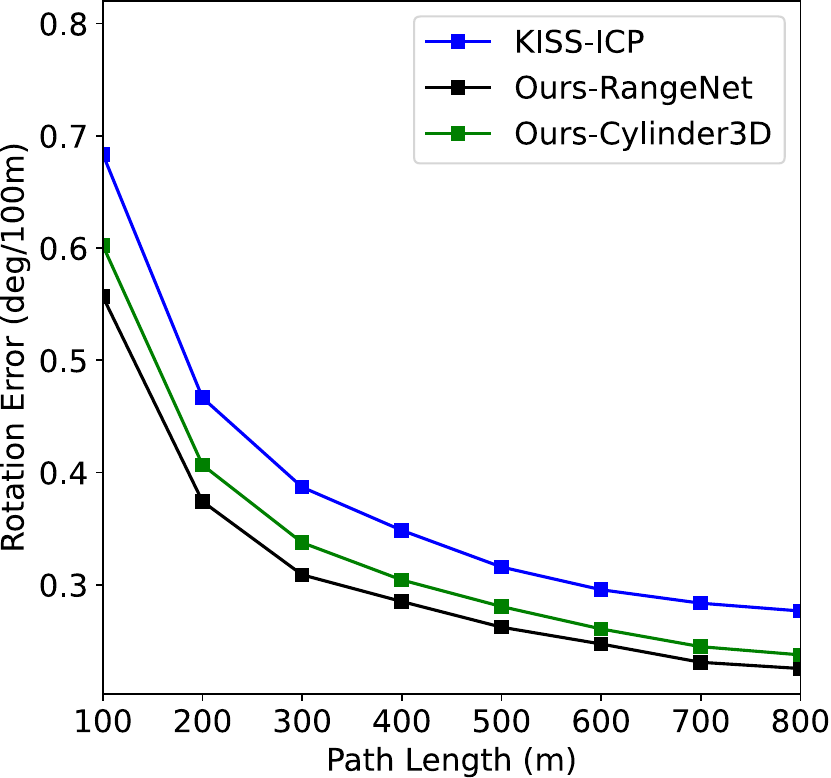} 
  \end{minipage}
}
\\
    \subcaptionbox{KITTI-360 t2}{
	\begin{minipage}[b]{0.5\textwidth}
\centering
\includegraphics[width=0.485\textwidth, height=0.4\textwidth]{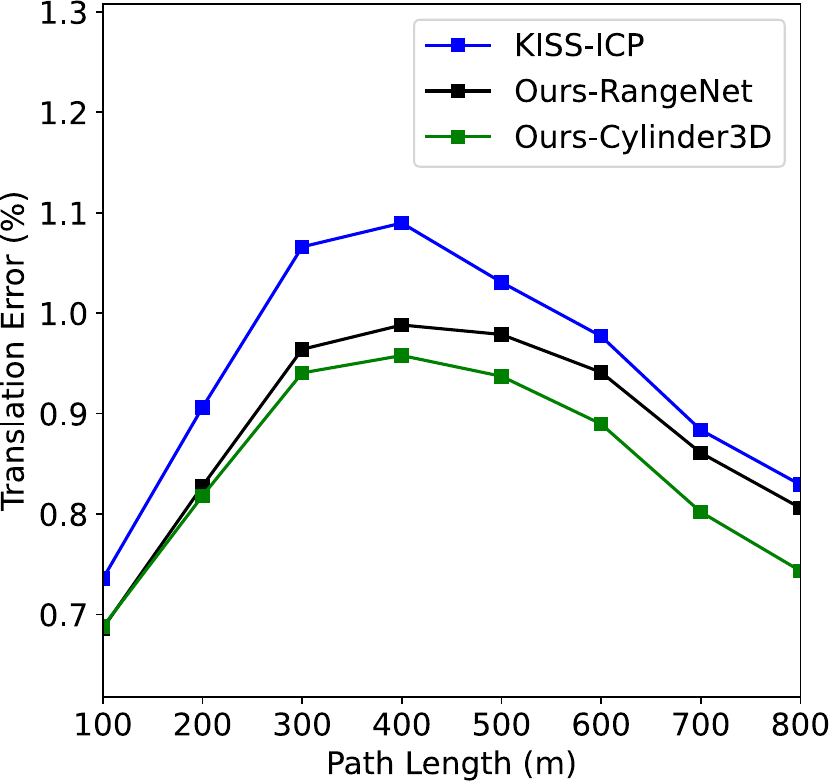}
\hspace{-7pt}
\includegraphics[width=0.487\textwidth, height=0.4\textwidth]{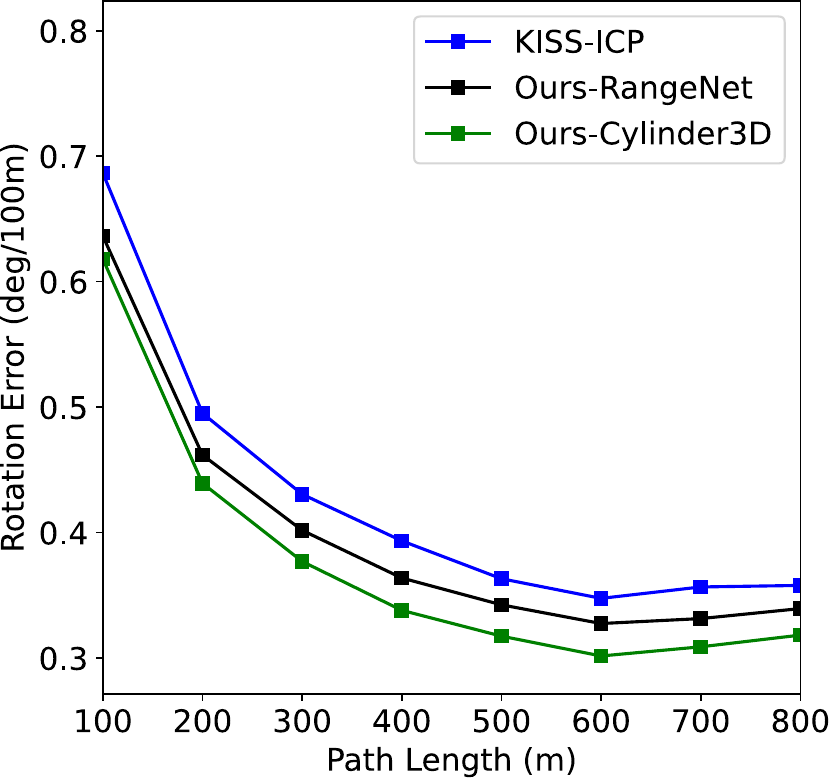} 
  \end{minipage}
}
    
    \caption{Relative pose errors versus length ranging from 100 to 800m. (a) The field of view is filled with a large number of buildings and intersecting roads. (b) Sequence t2 is similar to KITTI raw data, in that only a small amount of feature information can be extracted in the environment.}
    \label{rpe}
\end{figure}

KITTI-360 has nine sequences that are much longer than KITTI. 
The testing SLAM sequences are fragments of origin sequence 08, and we renamed them into sequences t0-t3, listed in KITTI-PART. Our method performs better or equally compared to the baseline methods in most sequences (see Table \ref{kitti} and Fig. \ref{rpe}) but fails in 03 and 10 due to many vans and trucks being incorrectly inferred into buildings. This error can be mitigated by improving the semantic segmentation model in future work.

    

\subsection{Real-time Performance and Ablation Studies}

Table \ref{time} presents the processing time of each part of our system. The semantic segmentation part can achieve real-time performance thanks to three independent processes (see Section \ref{seg}). The frame rate of the registration section also exceeds that of LiDAR, by about 11.60 Hz. These results show that our system is ready for real-world applications.

To verify the positive effect of each part of our system on improving the accuracy of pose estimation, we removed parts removing dynamic instance (RDI, Section \ref{rdi}), semantic subsampling (SS, Section \ref{ss}), semantic assisted association (SAA, Section \ref{saa}) and adaptive voxel map (AVM, Section \ref{avm}) respectively. We present the testing results of sequence t2 in Table \ref{ab} to verify the effectiveness of each module. Note that removing the semantic subsampling part can even lead to increased error, because many tiny semantic key features are filtered out by the traditional downsampling method, leaving many isolated noise-like points challenging for semantic-based point-pair association.

\begin{table}[h]
\caption{Ablation Studies}
\centering
\label{ab}
\begin{tabular}{c|c|c}
\toprule
&RTE ($\%$)&RRE (deg) \\
\midrule
Base&0.9392&0.44 \\
Ours-wo-RDI&0.8818&0.41 \\
Ours-wo-SS&0.9627&0.45 \\
Ours-wo-SAA&0.8537&0.40 \\
Ours-wo-AVM&0.8680&0.41 \\
Ours&\pmb{0.8467}&\pmb{0.39} \\
\bottomrule
\end{tabular}

\end{table}

\section{Conclusions}

This paper presents a semantic information-assisted ICP. We construct an online pose estimation system integrating pure LiDAR semantic information. Evaluation of the proposed method on KITTI Odometry Benchmark, KITTI-360, and specifically KITTI road sequences demonstrates that our approach can improve the localization accuracy in dynamic scenes with fewer effective geometric features. In future work, we plan to explore the fusion of semantic information in loop closure detection and incorporate other sensors to further improve semantic segmentation and pose estimation.

\addtolength{\textheight}{-5cm}   
\bibliographystyle{ieeetr}
\bibliography{root}

\end{document}